\pdfoutput=1

\documentclass[11pt]{article}

\usepackage{EMNLP2023}

\usepackage{cite}
\usepackage{amsmath,amssymb,amsfonts}
\usepackage{algorithmic}
\usepackage{graphicx}
\usepackage{textcomp}
\usepackage{xcolor}

\usepackage[group-separator={,}]{siunitx}
\def\BibTeX{{\rm B\kern-.05em{\sc i\kern-.025em b}\kern-.08em
    T\kern-.1667em\lower.7ex\hbox{E}\kern-.125emX}}

\usepackage{booktabs} 

\usepackage{times}
\usepackage{latexsym}

\usepackage[T1]{fontenc}

\usepackage[utf8]{inputenc}

\usepackage{microtype}

\usepackage{inconsolata}

\definecolor{red}{RGB}{255,0,0}
\definecolor{green}{RGB}{0, 102, 0}
\definecolor{maron}{RGB}{102,51,0}
\definecolor{orange}{RGB}{255,128,0}
\definecolor{blue}{RGB}{0,153,153}

\newcommand{\guillemet}[1]{``#1''}

\title{Deepparse : An Extendable, and Fine-Tunable State-Of-The-Art Library for Parsing Multinational Street Addresses}

\author{David Beauchemin, Marouane Yassine\\
Department of Computer Science and Software Engineering, Laval University \\
Group for Research in Artificial Intelligence of Laval University (GRAIL)\\
Québec, Canada \\
david.beauchemin@ift.ulaval.ca,
marouane.yassine.1@ulaval.ca\\
}

\DeclareFixedFont{\ttb}{T1}{txtt}{bx}{n}{12} 
\DeclareFixedFont{\ttm}{T1}{txtt}{m}{n}{12}  

\usepackage{color}
\definecolor{deepblue}{rgb}{0,0,0.5}
\definecolor{deepred}{rgb}{0.6,0,0}
\definecolor{deepgreen}{rgb}{0,0.5,0}

\usepackage{listings}

\newcommand\pythonstyle{\lstset{
language=Python,
basicstyle=\ttm,
morekeywords={self},              
keywordstyle=\ttb\color{deepblue},
emph={MyClass,__init__},          
emphstyle=\ttb\color{deepred},    
stringstyle=\color{deepgreen},
frame=tb,                         
showstringspaces=false,
basicstyle=\small
}}

\lstnewenvironment{python}[1][]
{
\pythonstyle
\lstset{#1}
}
{}

\newcommand\pythoninline[1]{{\pythonstyle\lstinline!#1!}}

\begin{document}

\maketitle

\begin{abstract}

Segmenting an address into meaningful components, also known as address parsing, is an essential step in many applications from record linkage to geocoding and package delivery. Consequently, a lot of work has been dedicated to develop accurate address parsing techniques, with machine learning and neural network methods leading the state-of-the-art scoreboard. However, most of the work on address parsing has been confined to academic endeavours with little availability of free and easy-to-use open-source solutions.

This paper presents Deepparse, a Python open-source, extendable, fine-tunable address parsing solution under LGPL-3.0 licence to parse multinational addresses using state-of-the-art deep learning algorithms and evaluated on over 60 countries. 
It can parse addresses written in any language and use any address standard. The pre-trained model achieves average $99~\%$ parsing accuracies on the countries used for training with no pre-processing nor post-processing needed. Moreover, the library supports fine-tuning with new data to generate a custom address parser.
\end{abstract}

\section{Introduction}
\emph{Address Parsing} is the task of decomposing an address into its different components \citep{8615844}. This task is essential to many applications, such as geocoding and record linkage. Indeed, it is quite useful to detect the different parts of an address to find a particular location based on textual data to make an informed decision. Similarly, comparing two addresses to decide whether two or more database entries refer to the same entity can prove to be quite difficult and prone to errors if based on methods such as edit distance algorithms given the various address writing standards.

There have been many efforts to solve the address parsing problem. From rule-based techniques \citep{rule-based} to probabilistic approaches and neural network models \citep{8615844}, much progress has been made in reaching accurate addresses segmentation. These previous works did a remarkable job of finding solutions for the challenges related to the address parsing task. However, most of these approaches either do not take into account parsing addresses from different countries or do so but at the cost of a considerable amount of meta-data and substantial data pre-processing pipelines \citep{rnn-parsing, hmm-parsing, crf-parsing, feedforward-parsing}.

However, most of the work on address parsing has been confined to academic endeavours with little availability of free and easy-to-use open-source solutions. In an effort to solve some of the limitations of previous methods, as well as offer an open-source address parsing solution, we have created \textbf{Deepparse}\footnote{\href{https://deepparse.org/}{https://deepparse.org/}} \citep{deepparse} an LGPL-3.0 licenced Python library. Our work allows anyone with a basic knowledge of Python or command line terminal 
to conveniently parse addresses from multiple countries using state-of-the-art deep learning models proposed by \citet{yassine2020leveraging, yassine2022multinational}. 
Deepparse's goal is to parse multinational addresses written in any language or using any address writing format with an extendable and fine-tunable address parser. In addition, \textbf{Deepparse} proposes a functionality to easily customize the aforementioned models to new data along with an easy-to-use Docker FastAPI
to parse addresses.

This paper's contributions are: First, we describe an open-source Python library for multinational address parsing. Second, we describe its implementation details and natural extensibility due to its fine-tuning possibilities. Third, we benchmark it against other open-source libraries.

\section{Related work}

Address parsing has been approached on the academic front using probabilistic machine learning models such as Hidden Markov Models and Conditional Random Fields (CRF) \citep{hmm-parsing, crf-parsing, 8615844}, as well as deep learning models mainly based on the recurrent neural network (RNN) architecture \citep{feedforward-parsing, rnn-parsing, 8615844}. Regarding openly available software, most of the existing packages cater to US postal addresses. For instance, \emph{pyaddress}\footnote{\href{https://github.com/SwoopSearch/pyaddress}{https://github.com/SwoopSearch/pyaddress}} allows for the decomposition of US addresses into eight different attributes with a possibility to specify acceptable \guillemet{street names}, \guillemet{cities} and \guillemet{street suffixes} in order to improve parsing accuracy. Similarly, \emph{address-parser}\footnote{\href{https://github.com/CivicKnowledge/address\_parser}{https://github.com/CivicKnowledge/address\_parser}} identifies as \guillemet{Yet another python address parser for US postal addresses} and enables users to extract multiple address components such as \guillemet{house numbers}, \guillemet{street names}, \guillemet{cardinal directions} and \guillemet{zip codes}. These two packages are based on a combination of predefined component lists and regular expressions. In contrast, \emph{usaddress}\footnote{\href{https://github.com/datamade/usaddress}{https://github.com/datamade/usaddress}} uses a probabilistic model that users can fine-tune using their data. 
Another openly available avenue for address parsing is Geocoding APIs, which can result in highly precise parsed addresses based on reverse geocoding. However, while being openly available, Geocoding APIs are often not free and not always convenient to use for a programming layperson.

The aforementioned approaches are limited to parsing addresses from a single country and either cannot handle a multinational scope of address parsing or would need to be adjusted to do so. To tackle this problem, Libpostal\footnote{\href{https://github.com/openvenues/libpostal}{https://github.com/openvenues/libpostal}}, a C library for international address parsing, has been proposed. This library uses a CRF-based model trained with an averaged Perceptron for scalability. The model was trained on Libpostal dataset\footnote{\href{https://github.com/openvenues/libpostal\#training-data}{https://github.com/openvenues/libpostal\#training-data}} and achieved a $99.45~\%$ full parse accuracy\footnote{The accuracy was computed considering the entire sequence and was not focused on individual tokens.} using an extensive pre and post-processing pipeline. 
However, this requires putting addresses through a heavy pre-processing pipeline before feeding them to the prediction model, and it does not seem possible to develop a new address parser based on the documentation. 
A thorough search of the relevant literature yielded no open-source neural network-based software for multinational address parsing.

\section{Implementation}
Deepparse is divided into three high-level components: pre-processors, embeddings model, and tagging model. 
The first component, the pre-processor, is a series of simple handcrafted pre-processing functions to be applied as a data cleaning procedure before the embedding component, such as lowercasing the address text and removing commas. By default, Deepparse simply lowercase and removes all commas in the address. The library does not require a complex pre-processing pipeline, but one can be defined and used more complex one if needed since Deepparse is built so users can handcraft and use a custom pre-processor during this phase.

\begin{figure*}
    \centering
    \includegraphics[scale=0.35]{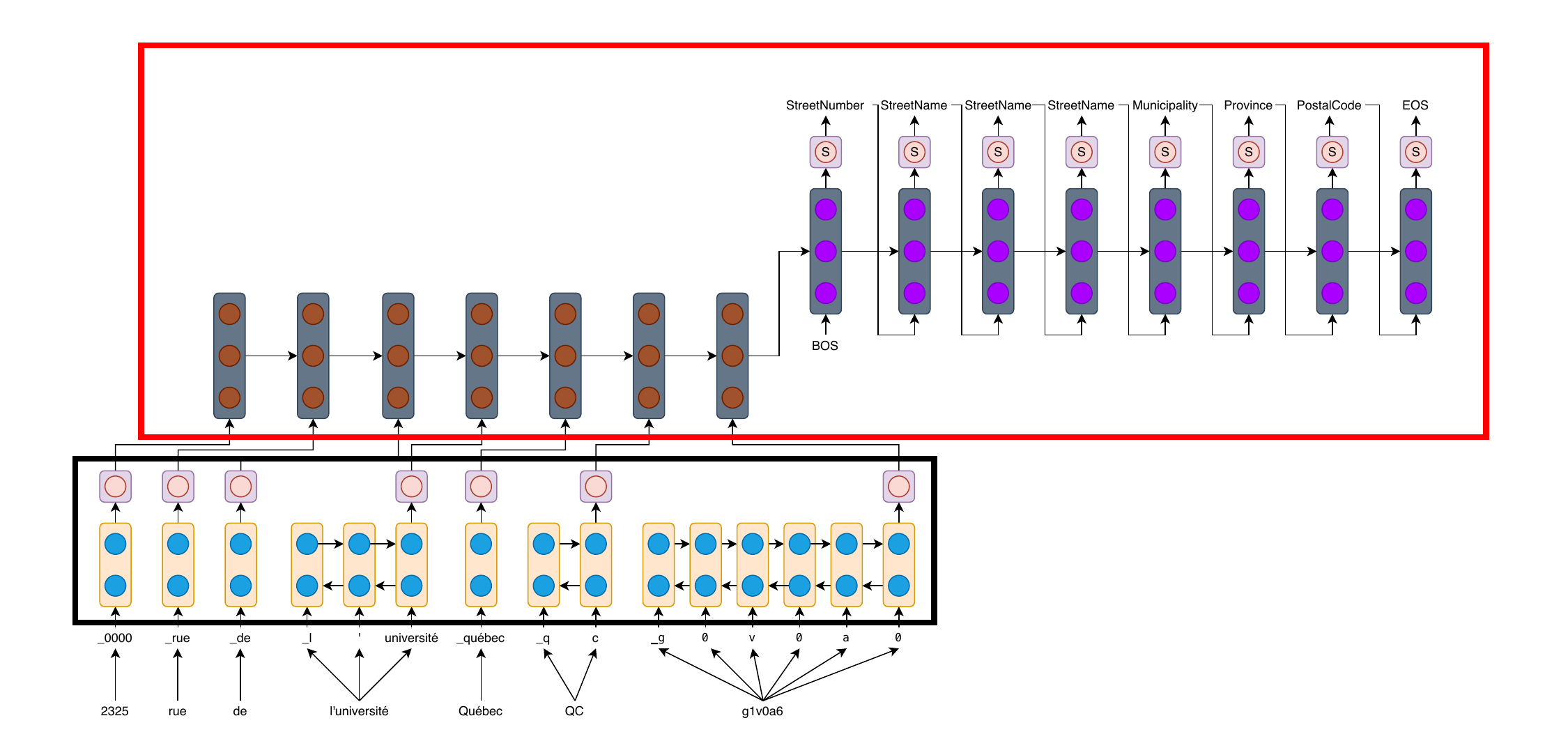}
    \caption{Illustration of our architecture using one of the two embedding model component (black) approach. Each word in the address is encoded using an embedding model, in this case, MultiBPEmb (the BPE segmentation algorithm replaces the numbers in the address with zeros). The embeddings are fed to a BiLSTM (rounded rectangle with two circles). The last hidden state for each word is run through a fully connected layer (rounded rectangle with one circle). The resulting embeddings are given as input to the tagging model components (red). The \guillemet{S} in the fully connected layer following the Seq2Seq decoder stands for the Softmax function.}
    \label{fig:my_label}
    \vspace{-1em}
\end{figure*}

The last two components are illustrated in \autoref{fig:my_label}. We can see that the embeddings model component (black) encodes each token (i.e. word) of the address into a recurrent dense representation. At the end of the sentence, the component generates a single dense representation for the overall address generated from the individual address components.
Then, this address-dense representation is used as input to the tagging model component (red), where each address component is decoded and classified into its appropriate tag.
These two components do not rely on named entity recognition to parse addresses as opposed to the one proposed by \citet{8615844}.

Deepparse proposes two embeddings model approaches and four pre-trained tagging model architectures; all approaches can be used with CPU or GPU setup. 
All pre-trained approaches have been trained on our publicly available dataset\footnote{\href{https://github.com/GRAAL-Research/deepparse-address-data}{https://github.com/GRAAL-Research/deepparse-address-data}}, based on to the Libpostal dataset, and achieved parse accuracies higher than 99\% on the 20 trained countries without using pre or post-processing\footnote{ The accuracy for each sequence is computed as the proportion of the tags predicted correctly by the model. Predicting all the tags correctly for a sequence yields perfect accuracy.}.

The following sub-section will briefly discuss how these two components work. For more details on the algorithms behind both components, readers can refer to \citet{yassine2020leveraging, yassine2022multinational}. 
We will finish this section with a presentation on Deepparse's ability to developing a new address parser.

\subsection{Embedding Model}
Our objective was to build a single neural network to parse addresses from multiple countries. Thus, access to embeddings for different languages at runtime was necessary. Since the use of alignment vectors \citep{fasttext-aligned, muse} would have introduced the unnecessary overhead of detecting of the source language to project word embeddings from different languages in the same space, Deepparse proposes the following two methods.

First, we use a fixed pre-trained monolingual French fastText model. We chose French embeddings since this language shares Latin roots with many languages in our test set. It is also due to the large corpus on which these embeddings were trained. We refer to this embeddings model technique as \textbf{fastText}.

Second, we use an encoding of words using MultiBPEmb and merge the obtained embeddings for each word into one word embedding using an RNN. This method has been shown to give good results in a multilingual setting \citep{heinzerling-strube-2019-sequence}. Our RNN network of choice is a Bidirectional LSTM (Bi-LSTM) with a hidden state dimension of 300. We build the word embeddings by running the concatenated forward and backward hidden states corresponding to the last time step for each word decomposition through a fully connected layer of which the number of neurons equals the dimension of the hidden states. This approach produces 300-dimensional word embeddings. We refer to this embeddings model technique as \textbf{BPEmb}. 

\subsection{Tagging Model}
\label{subsec:tagging}
Our downstream tagging model is a Seq2Seq model. 
Using Seq2Seq architecture as tagging model is effective for data with sequential pattern \citep{huang2019tagging, omelianchuk2021text, jin2021hierarchical, raman2022transforming} such as address.
The architecture consists of a one-layer unidirectional LSTM encoder and a one-layer unidirectional LSTM decoder followed by a fully-connected linear layer with a softmax activation. Both the encoder's and decoder's hidden states are of dimension $1024$.
The embedded address sequence is fed to the encoder that produces hidden states, the last of which is used as a context vector to initialize the decoder's hidden states. The decoder is then given a \guillemet{Beginning Of Sequence} (BOS) token as input, and at each time step, the prediction from the last step is used as input. To better adapt the model to the task at hand and to facilitate the convergence process, we only require the decoder to produce a sequence with the same length as the input address. This approach differs from the traditional Seq2Seq architecture in which the decoder makes predictions until it predicts the ends-of-sequence token. The decoder's outputs are forwarded to the linear layer, of which the number of neurons equals the tag space dimensionality. The softmax activation function computes probabilities over the linear layer's outputs to predict the most likely token at each time step.

Deepparse proposes four pre-trained tagging model architectures: one using each embedding model approach, namely \textbf{fastText} and \textbf{BPEmb}, and one using each embedding model approach with an added attention mechanisms. Attention mechanisms are neural network components that can produce a distribution describing the interdependence between a model's inputs and outputs (general attention) or amongst model inputs themselves (self-attention). These mechanisms are common in natural language processing encoder-decoder architectures such as neural machine translation models \citep{nmt} since they have been shown to improve models' performance and help address some of the issues RNNs suffer from when dealing with long sequences. Also, \citet{yassine2022multinational} has shown that the attention mechanism has significantly increased performance for incomplete addresses. Incomplete addresses do not include all the components defined by a country-written standard—for example, an address missing its postal code. They are cumbersome and cause problems for many industries, such as delivery services and insurance companies \citep{nagabhushan2009soft}.

\paragraph{Choosing a Model}
The difference between all four models is their capabilities to generate better results on unseen address patterns and unseen language.
For example, as shown in \citet{yassine2020leveraging}, BPEmb embeddings models generate better parsing on address from India, even if the language and address pattern was unseen during training compared to FastText embeddings model.
However, this increase in generalization performance comes at the cost of longer inference time (will be discussed in \autoref{sec:pratical}).
As shown in \citet{yassine2022multinational}, models using the attention mechanism also demonstrate the same improved generalization performance compared to their respective embeddings approaches but with the same cost of inference performance.
Thus, one must trade off generalization performance over inference performance.


\subsection{Developing a New Parser}
One of the unique particularities of Deepparse is the ability to develop a new parser for one's specific needs. Namely, one can fine-tune one of our pre-trained models for their specific needs using our public dataset or theirs. 
Doing so can improve Deepparse's performance on new data or unseen countries, giving Deepparse great flexibility.
As shown in \autoref{fig:codeexample1}, developing (i.e. fine-tuning) a new parser using our pre-trained public models is relatively easy and can be done with a few Python lines of code.

\begin{figure*}
    \centering
    \begin{python}
address_parser = AddressParser(model_type="fasttext")
address_parser.retrain(dataset, train_ratio=0.8, epochs=5)
    \end{python}
    \caption{Code example to fine-tune our \texttt{"FastText"} pre-trained model on a new \texttt{dataset} for $5$ epochs using a 80-20~\% train-evaluation dataset ratio.}
    \label{fig:codeexample1}
\end{figure*}

Moreover, as shown in \autoref{fig:codeexample2}, one can also use Deepparse to retrain our pre-trained models on new prediction tags easily, and it is not restricted to the ones we have used during training, making it flexible for new addresses pattern.

\begin{figure*}
    \centering
    \begin{python}
address_parser = AddressParser(model_type="fasttext")
new_tag_dictionary = {"ATag": 0, "AnotherTag": 1, "EOS": 2}
address_parser.retrain(dataset, prediction_tags=tag_dictionary)
    \end{python}
    \caption{Code example to retrained our \texttt{"FastText"} pre-trained model on a new \texttt{dataset} with new \texttt{tags}.}
    \label{fig:codeexample2}
\end{figure*}

Finally, as shown in \autoref{fig:codeexample3} it is also possible to easily reconfigure the tagging model architecture to either create a smaller architecture, thus potentially reducing memory usage and inference time, or increase it to improve performance on more complex address data. Also, one can do all of the above at the same time.

\begin{figure*}
    \centering
    \begin{python}
address_parser = AddressParser(model_type="fasttext")
seq2seq_params = { "encoder_hidden_size": 512, "decoder_hidden_size": 512}
address_parser.retrain(dataset, seq2seq_params=seq2seq_params)
    \end{python}
    \caption{Code example to train a new model using our Seq2Seq architecture with a different configuration (i.e. encoder and decoder hidden size).}
    \label{fig:codeexample3}
\end{figure*}

\section{Practical results}
\label{sec:pratical}

In this section, since Libpostal and Deeparse are comparable in terms of accuracy, both are almost perfect; we benchmark Deepparse memory usage and inference time with 183,000 addresses of the Deepparse dataset.
Our parsing experiment processes 183,000 addresses using different batch sizes ($2^0, \dots, 2^9$) and assesses memory usage and inference time performance for Libpostal and Deepparse. 
Since Deepparse can batch address, we assess the inference time as the average processing time per address (i.e. $\frac{\text{Total time to process all addresses}}{183,000} = \text{time per address}$). Libpostal does not offer batching functionality. The experiment used a GPU and a CPU to assess the accelerator's gain. Thus, we also assess GPU memory usage in our experiment that uses such devices.

Our experiment was conducted on Linux OS 22.04, with the latest Python version (i.e. 3.11), Python memory\_profiler 0.61.0, Torch 2.0 and CUDA 11.7 (done March 21, 2023). Our GPU device is an RTX 2080.

\autoref{tab:gpu} and \autoref{tab:cpu} present our experiment results using respectively a GPU device or not with or without using batch processing. 
In both tables, we can see that Libpostal achieved better inference time performance. However, Deepparse still achieved interesting performance, particularly with batching that reduced by one order of magnitude the average processing time of execution.

\begin{table}
    \resizebox{0.5\textwidth}{!}{%
    \begin{tabular}{lcccc}
    \toprule
     & \textbf{\begin{tabular}[c]{@{}c@{}}GPU\\Memory usage\\(GB)\end{tabular}} & \textbf{\begin{tabular}[c]{@{}c@{}}RAM\\usage\\(GB)\end{tabular}} & \textbf{\begin{tabular}[c]{@{}c@{}}Mean time\\of execution\\ (not batched) (s)\end{tabular}} & \textbf{\begin{tabular}[c]{@{}c@{}}Mean time\\of execution\\ (batched) (s)\end{tabular}} \\\midrule
    \textbf{fastText} & $\sim$1 & $\sim$8 & $\sim$0.0023 & $\sim$0.0004 \\
    \textbf{fastTextAttention} & $\sim$1.1 & $\sim$8 & $\sim$0.0043 & $\sim$0.0007 \\
    \textbf{BPEmb} & $\sim$1 & $\sim$1 & $\sim$0.0055 & $\sim$0.0015 \\
    \textbf{BPEmbAttention} & $\sim$1.1 & $\sim$1 & $\sim$0.0081 & $\sim$0.0019 \\\midrule
    Libpostal & 0 & $\sim$2.3 & $\sim$0.00004 & $\sim$N/A\\\bottomrule
    \end{tabular}%
    }
    \caption{GPU and RAM usage and average processing time to parse 183,000 addresses using a GPU device with or without batching.}
    \label{tab:gpu}
    \vspace{-1em}
\end{table}

\begin{table}
    \resizebox{0.5\textwidth}{!}{%
    \begin{tabular}{lcccc}
    \toprule
     & \textbf{\begin{tabular}[c]{@{}c@{}}RAM\\usage\\(GB)\end{tabular}} & \textbf{\begin{tabular}[c]{@{}c@{}}Mean time\\of execution\\ (not batched) (s)\end{tabular}} & \textbf{\begin{tabular}[c]{@{}c@{}}Mean time\\of execution\\ (batched) (s)\end{tabular}} \\\midrule
    \textbf{fastText} & $\sim$8 & $\sim$0.0128 & $\sim$0.0026 \\
    \textbf{fastTextAttention} & $\sim$8 & $\sim$0.0230 & $\sim$0.0057 \\
    \textbf{BPEmb} & $\sim$1 & $\sim$0.0179 & $\sim$0.0044 \\
    \textbf{BPEmbAttention} & $\sim$1 & $\sim$0.0286 & $\sim$0.0075 \\\midrule
    Libpostal & $\sim$1 & $\sim$0.00004 & $\sim$N/A\\\bottomrule
    \end{tabular}%
    }
    \caption{RAM usage and average processing time to parse 183,000 addresses using only CPU with or without batching.}
    \label{tab:cpu}
    \vspace{-1em}
\end{table}

\section{Future Development and Maintaining the Library}

As our development roadmap, we plan to improve the documentation by adding a training guide on how one can develop its address parser.
Also, we plan to offer new deep learning architecture that leverages more recent progress, such as a Transformer based architecture and to support more words embedding models, such as contextualized embeddings like ELMO embeddings \citep{peters2018deep}. 
Moreover, we plan to offer a minimalist application to address parsing for coding laypersons.
Finally, we aim at improving inference time performance by using recent integration of quantization technique \citep{cheng2018recent, wu2020integer} in PyTorch, namely, \guillemet{performing computations and storing tensors at lower bitwidths than floating point precision} \citep{pytorch2}.
The library is maintained mainly by the library authors, and three to four releases are published yearly to improve and maintain the solution.

\section{Conclusion}
In conclusion, we have described Deepparse, an extendable and fine-tunable state-of-the-art library for parsing multinational street addresses. It is an open-source library, has over 99.9\% test coverage and integrates easily with existing natural language processing pipelines. Deepparse offers great flexibility to users who can develop their address parser using our easy-to-use fine-tuning interface.
Although slower than the Libpostal alternative implemented in low-level language C, Deepparse successfully parses more than 99\% of address components. 

\section*{Acknowledgment}
This research was supported by the Natural Sciences and Engineering Research Council of Canada (IRCPJ 529529-17) and a Canadian insurance company. 
We wish to thank the reviewers for their comments regarding our work and methodology.

\bibliographystyle{acl_natbib}
\bibliography{nlposs2023}

\vspace{12pt}

\end{document}